\documentclass[11pt,a4paper]{article}
\usepackage[hyperref]{acl2017}
\usepackage{times}
\usepackage{latexsym}
\usepackage{graphicx}
\usepackage{float}

\usepackage{url}

\aclfinalcopy % Uncomment this line for the final submission
\newcommand{\generictitle}{Argument Labeling of \protect\\ Explicit Discourse Relations using \protect\\ LSTM Neural Networks\\}
\newcommand{\addr}{Depart. of Computer Science and Software Engineering\\
  Concordia University\\
  1515 Ste-Catherine Street West\\
  Montr{\'e}al, Qu{\'e}bec, Canada H3G 2W1\\ }
  
\title{\generictitle}

\author{Sohail Hooda \hspace{2cm} Leila Kosseim\\
  \addr
  {\tt \{s\_hooda, kosseim\}@encs.concordia.ca} \\
}

\date{}

\begin{document}
\maketitle
\begin{abstract}
  Argument labeling of explicit discourse relations is a challenging task. The state of the art systems achieve slightly above 55\% F-measure but require hand-crafted features. In this paper, we propose a Long Short Term Memory (LSTM) based model for argument labeling. We experimented with multiple configurations of our model. Using the PDTB dataset, our best model achieved an F1 measure of 23.05\% without any feature engineering. This is significantly higher than the 20.52\% achieved by the state of the art RNN approach, but significantly lower than the feature based state of the art systems. On the other hand, because our approach learns only from the raw dataset, it is more widely applicable to multiple textual genres and languages.  
\end{abstract}

\section{Introduction}
\label{sec:intro}
    In well written texts, discourse relations are used to provide additional meaning to the underlying content by connecting two textual segments logically. This in turn facilitates the reader's understanding of the text. For example, in:

%    \begin{exe}
%	    \ex
    \label{ex:one} (1) \textit{We would stop index arbitrage} \underline{when} \textbf{the market is under stress}. \footnote{\label{note1} This example is taken from the Penn Discourse Treebank \citep{PDTB2Annotation}. }    
%   \end{exe}

    two discourse segments, or arguments (\texttt{Arg1} in italics and \texttt{Arg2} in bold), are explicitly connected via the \underline{connective} underlined and related by the discourse relation of \textsc{Condition}. Discourse relations can be made explicit through the use of discourse connectives such as \textit{although}, \textit{but}, \textit{since}, \textit{because}, etc. or can be left implicit, when no explicit cue phrase is used to signal the relation. 
    
    Discourse parsing involves two main tasks: (1) argument labeling, or identifying the boundaries and labeling \texttt{Arg1} and \texttt{Arg2}, and (2) relation labeling, or identifying the discourse relation that holds between the arguments. Because discourse parsing allows a deeper understanding of the communicative goal of text segments, it has been used in a variety of downstream NLP applications such as text summarization \citep{barzilay2004catching, yoshida2014dependency} and question-answering \citep{chai2004discourse, verberne2007evaluating}.
    
    As witnessed in the recent CoNLL shared tasks \cite{conll2015, conll2016}, full end-to-end discourse parsing is still a challenge. In particular, argument labeling is difficult as the exact boundaries of both \texttt{Arg1} and \texttt{Arg2} must be identified. The state-of-the-art system \citep{wang2016two} achieves an F-measure of only 55.11\% and thus leaves much room for improvement. Most work in this domain make use of a variety of hand-crafted features that do not handle long-distance dependencies well. However, the great majority of discourse arguments are not adjacent to one another and long distance features are important for this task. 
    
    In this paper, we investigate the use of recurrent neural networks for discourse labeling. Specifically, we investigate the use of Long Short Term Memory (LSTMs) to better handle long term dependencies and automatically extract and embed the features without prior input. We show that a widely applicable model can be produced by learning from the input data alone, and learning the features directly. This allows more generalized applications of our model across multiple text genres as well as languages. We also show that the approach does not suffer from long distance dependencies and achieves stable results regardless of the distance between \texttt{Arg1} and \texttt{Arg2}. To our knowledge, this is the first attempt at argument labelling using Deep Learning architectures that uses no hand-crafted  features and achieves good results in contrast to the existing systems which rely on hand-crafted features during the learning process of the model.

\section{Related Work}
\label{sec:related-work}
    Due to the CoNLL 2015 and 2016 shared tasks \citep{conll2015, conll2016}, much recent work has addressed the problem of discourse parsing. However, much work is still needed in order to reach the human performance of 82.8\% reported by \citep{miltsakaki2004annotating}. Discourse parsing consists of both: (1) argument labeling and (2) relation labeling (e.g. \citep{lin2014pdtb, laali2015clac, kong2014constituent, conll2015, conll2016}).  The approaches used for argument labeling at CoNLL 2015 (e.g. \citep{laali2015clac, son2015jaist, zhou2015sonlp} largely consisted of standard supervised machine learning techniques such as Support Vector Machines (SVMs), Conditional Random Fields, Naive Bayes and Maximum Entropy (MaxEnt) and viewed the problem as a sequence labeling task. These models used syntactic positional and lexical features (such as the first word after the discourse connective) to learn to identify discourse arguments. These traditional methods achieved F-scores in the order of 45-55\%. In 2016, several Convolutional Neural Networks (CNNs) and Recurrent Neural Networks (RNNs) were introduced for relation labeling \citep{qin2016shallow}. %eisenstein2016shallow}.
    However, for argument labeling, only \citep{wang2015dcu} used RNNs with word embeddings as input which were learned specifically from the training corpus and combined these with hand engineered features based on part of speech tags as well as other linguistic information. The group also used a classifier to determine whether a discourse relation spanned over more than one sentence and based on its output used separate classifiers for same-sentence and multiple-sentence relations. However their F1 score of 20.52\% was still significantly below the F1 scores of traditional methods. 
    While applying hand engineered features in a machine learning model does provide good results, it however, forces the model to be specific to the dataset that it is applied to. As a result, the model is unable to perform well with other datasets where either the content of the dataset differs or the language \citep{prasad2011biomedical}. In contrast, if a machine learning model is configured to learn features solely on the training dataset, it would be able to generalize over larger datasets of different domain with relative ease.

\section{Motivation}
\label{sec:motivation}

    A major problem in argument labeling is that arguments may be dependent on long distance features. For example, in:
    
%    \begin{exe}
%	    \ex 
    \label{ex:two} (2) \textit{These are all market excesses} (putting aside the artificial boosts that the tax code gives to debt over equity), \underline{and} \textbf{what we've seen is the market reining them in}.
	    %\textit{Arg1}, \underline{Connective}, \textbf{Arg2}.
%    \end{exe}
    
    \texttt{Arg1} and \texttt{Arg2} are separated by 14 words\footnote{The connective acts as a marker for the start of \texttt{Arg2}, and therefore it is not included in this calculation.}. These words, known as {\em attribution}, are a challenge for standard features. For example, features from \citep{wang2015dcu} such as ``1\textsuperscript{st} Previous Word of Connective'', ``2\textsuperscript{nd} Previous Word of Connective'', ``1\textsuperscript{st} Previous POS of Connective '' or ``2\textsuperscript{nd} Previous POS of Connective'', do not handle non-adjacent arguments well. Table~\ref{tab:arg1-location-stats} shows the number of instances in the PDTB dataset \citep{PDTB2Annotation} used at CoNLL 2015 and 2016 that do not have consecutive arguments. As Table~\ref{tab:arg1-location-stats} shows, 78\% of \texttt{Arg1} and \texttt{Arg2} are not located consecutively and 24\% are separated by at least 5 words. Standard machine learning techniques have much difficulty to account for these and a standard RNN approach (as used by \cite{wang2015dcu}) also suffers from long distance dependencies between \texttt{Arg1} and \texttt{Arg2} due to the problem of vanishing/exploding gradients \citep{hochreiter2001gradient}. When attempting to incorporate information between words in arguments that are non-consecutive, as the distance between argument increases, the information collected either converges to zero or to infinity resulting in over- or under-estimating the learning of the neural network. On the other hand, LSTMs can control this behaviour and have been shown to model long distance dependencies much better \citep{hochreiter1997long}. For this reason, we experimented with such an approach.

\begin{table}[]
    \centering
    \caption{Statistics on the distance (in number of word tokens) between \texttt{Arg1} and \texttt{Arg2} in the explicit relations in the training and test set of the PDTB Dataset \citep{PDTB2.0}}
        \begin{tabular}{|c|r|r|r|}
            \hline
             \textbf{Distance} &    \textbf{Number of instances} &  \textbf{Percentage} \\\hline
             0              &   3,554   &   22.29\% \\
             1              &   8,582   &   53.82\% \\
             2-10           &   1,743   &   10.93\% \\
             \textgreater10 &   2,066   &   12.96\% \\ \hline
             \textbf{Total} &  \textbf{15,945}   &  \textbf{100.00}\% \\ \hline
        \end{tabular}
    \label{tab:arg1-location-stats}
    \end{table}

\section{Experiment}
\label{sec:experiment}

    \begin{figure*}[h!]
        %\centering
        \includegraphics[width=0.675\textwidth]{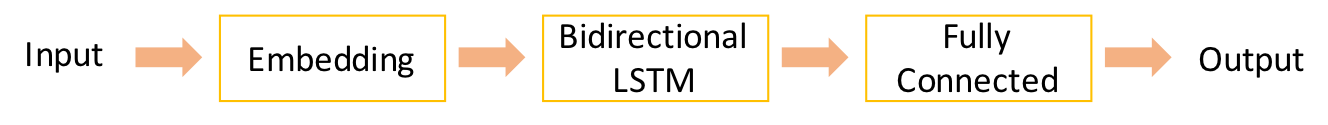} \\
        \includegraphics[width=1.0\textwidth]{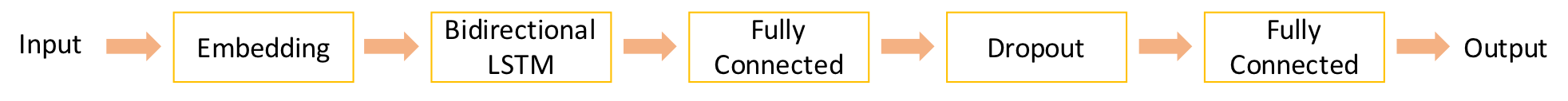}
        \caption{Architecture of model \texttt{m1} (top) and model \texttt{m2} (bottom)}
        \label{fig:exp}
    \end{figure*}

\subsection{Corpus}
\label{sec:corpus}
To train and validate our LSTM approach, we used the Penn Discourse Treebank Corpus (PDTB) \citep{PDTB2Annotation}. The PDTB has become the standard dataset in discourse parsing, thanks, in part, to the CoNLL shared tasks \citep{conll2015, conll2016}. Using this corpus allowed us to compare our work with the state of the art systems. The PDTB contains both explicit relations (marked with discourse connectives such as \textit{because} or \textit{but}) as well as non-explicit relations. Table~\ref{tab:data-stats} shows statistics of the dataset.

    \begin{figure*}[h!]
        %\centering
        \includegraphics[width=0.33\textwidth]{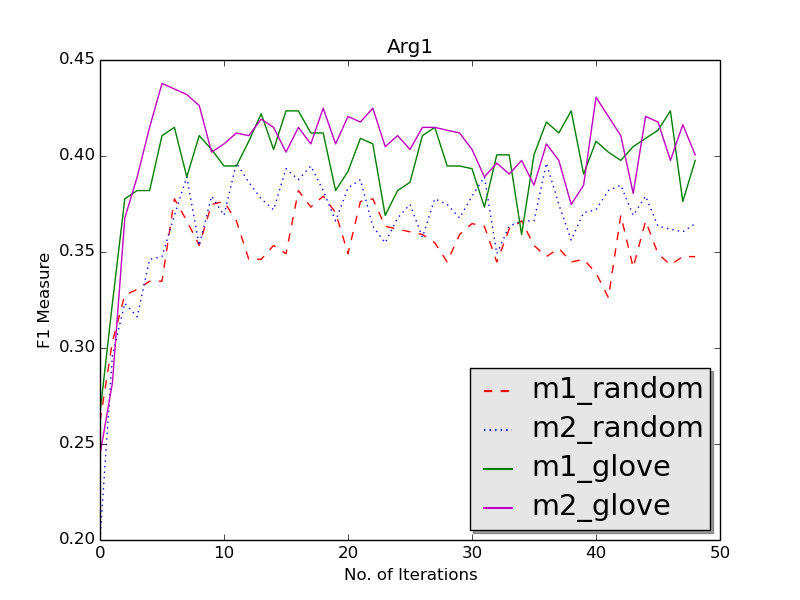}
        \includegraphics[width=0.33\textwidth]{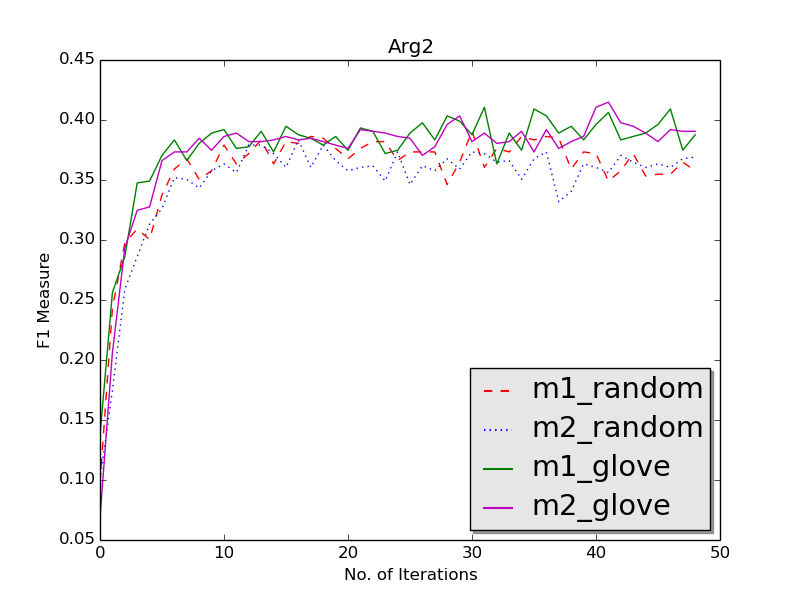}
        \includegraphics[width=0.33\textwidth]{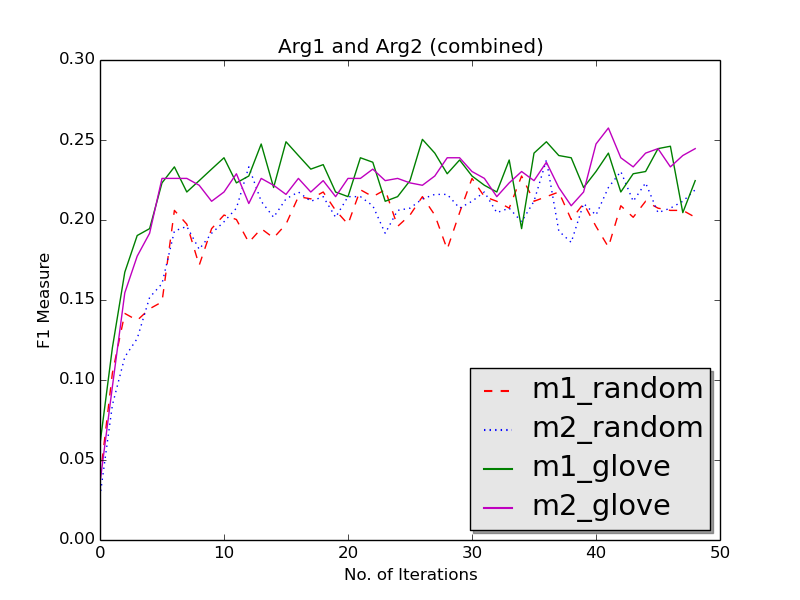}
        \caption{F1 score on the test set as a function of the number of iterations on the training set for \texttt{Arg1} (top), \texttt{Arg2} (middle) and \texttt{Arg1+Arg2} (bottom)}
        \label{fig:args}
    \end{figure*}
    
    \begin{table}[]
    \centering
    \caption{Number of instances in the PDTB Dataset}
        \begin{tabular}{|l|r|r|r|}
            \hline
             \textbf{Dataset} & \textbf{Explicit} & \textbf{Non-Explicit} & \textbf{Total} \\ \hline
             Training   & 15,246 & 17,289 & 32,535 \\
             Testing    &    699 &    737 &  1,436 \\ \hline
             \textbf{Total}      & \textbf{15,945} & \textbf{18,026} & \textbf{33,971} \\ \hline
        \end{tabular}
    \label{tab:data-stats}
    \end{table}
    
    Since we focused on explicit relations only, the dataset was first cleaned by removing all non-explicit relations. As is standard in the field, we used sections 2-21 of the PDTB for training and section 22 was used for testing. Thus, about 15,246 instances such as example~(\ref{ex:one}) in Section~\ref{sec:intro} were used for the training process and 699 were used for testing. 

    \begin{figure*}[h!] 
    \begin{center}
    \begin{tabular}{|l|}\hline \\
        %\centering
	        \textit{We would have to wait} \underline{until} \textbf{we have collected on} \textbf{those assets}\\ 
	    
	       [1,~~~2,~~~~~~~3,~~~~~~4,~5,~~~~~6,~~~~~~1,~~7,~~~~~~8,~~~~~~~~~~~~~9,~10,~~~~~11,~~~~~~0,~0,...,~0] \\ \\
	        \hline 
	        \end{tabular}
	        \end{center}
%    \end{exe}
        \caption{Example with words in a training instance labelled with the corresponding numeric value}
        \label{fig:instance}
    \end{figure*}
    
\subsection{Network Architecture}
\label{subsec:network}
For our experiments, we used a neural network composed of Long Short-Term Memory (LSTM) cells \citep{hochreiter1997long}. An LSTM is a specialized form of a Recurrent Neural Network (RNN) where a neuron is replaced with a memory cell. The memory cell is able to learn and hold information to take into account long term dependencies, thus allowing to overcome the problem of vanishing and exploding gradients with RNNs \citep{hochreiter2001gradient}.

    In order to learn the position of \texttt{Arg1} and \texttt{Arg2} and the length of these segments from the data only, we experimented with two main architectures. The first architecture shown in Figure~\ref{fig:exp} (top) is composed of an embedding layer that feeds directly into a Bidirectional LSTM layer. The Bidirectional LSTM layer is composed of 100 LSTM cells (for each direction) and the initialization is performed via the Glorot Uniform technique~\cite{Glorot2010}. The Bidirectional LSTM outputs are then fed into a fully connected layer which outputs the probability of 4 possible labels: \texttt{Arg1}, \texttt{Arg2}, \texttt{connective} or \texttt{none} for each input word. In the second architecture, shown in Figure~\ref{fig:exp} (bottom), we added a dropout layer as well as a fully connected layer at the end of the first model. This is shown in Figure~\ref{fig:exp}. We tested our architectures with different dimensions of word embeddings and decided to use a value of 300 as it resulted in a higher accuracy with the training set (see Section~\ref{sec:data-prep}). Thus, this created the 2 models below: 
    \begin{enumerate}
        \item {\tt m1}: Bidirectional LSTM layer  + vectors of 300 words 
        \item {\tt m2}: Bidirectional LSTM layers + Dense + Dropout + Dense + vectors of 300 words
    \end{enumerate}

    Using both models, we also experimented with randomly generated embeddings as well as pre-computed word embeddings from \citep{pennington2014glove} (see Section~\ref{sec:data-prep}). All models were learned over 50 epochs. The cost function was minimized via the Adam Optimizer \citep{kingma2014adam}, a memory efficient stochastic optimizer that relies on first order differentials only and calculates updates via the first and second order moments of the gradients thereby resulting in a linear update process. The cost was minimized over the mean of the labels for an entire mini batch provided in a single iteration.

\vspace*{-.15cm}
\subsection{Data Preparation}
\vspace*{-.15cm}

\label{sec:data-prep}
    To provide as input to the neural networks, the training instances were converted into a numeric matrix structure. Since word embeddings are updated dynamically, we used pre-computed \texttt{GloVe} embeddings (Global Vectors for Word Representation) \citep{pennington2014glove} in one set of experiments and random values in another set. This was done by creating a dictionary of words and assigning each word to a random numeric value. As shown in Figure~\ref{fig:instance}, a ``zero word'' was added to the vocabulary as a placeholder to pad sentences to an equal length. This length was set to 1,170 words, which is the size of the longest discourse segment containing both \texttt{Arg1} and \texttt{Arg2} segments in the PDTB training dataset. Thus the input data was a 2 dimensional matrix of a fixed size of 300 by 1,170. The use of fixed size vectors was not a necessary requirement for the network, but it was mechanically easier to have consistency within the dataset. The label vectors were also correspondingly padded with the \texttt{none} class. This allowed the network to learn the end of the \texttt{Arg1+Arg2} sequence.
%    \begin{exe}
%	    \ex
%    \label{ex:three}
    
    %The integers representing a word in the vocabulary were then replaced with a vector of size 128 or 300 depending on the experiment.
    
\section{Results and Analysis}
\label{sec:results-analysis}

\begin{figure*}[h!]
        \centering
        \includegraphics[width=1.1\textwidth,height=8cm,keepaspectratio]{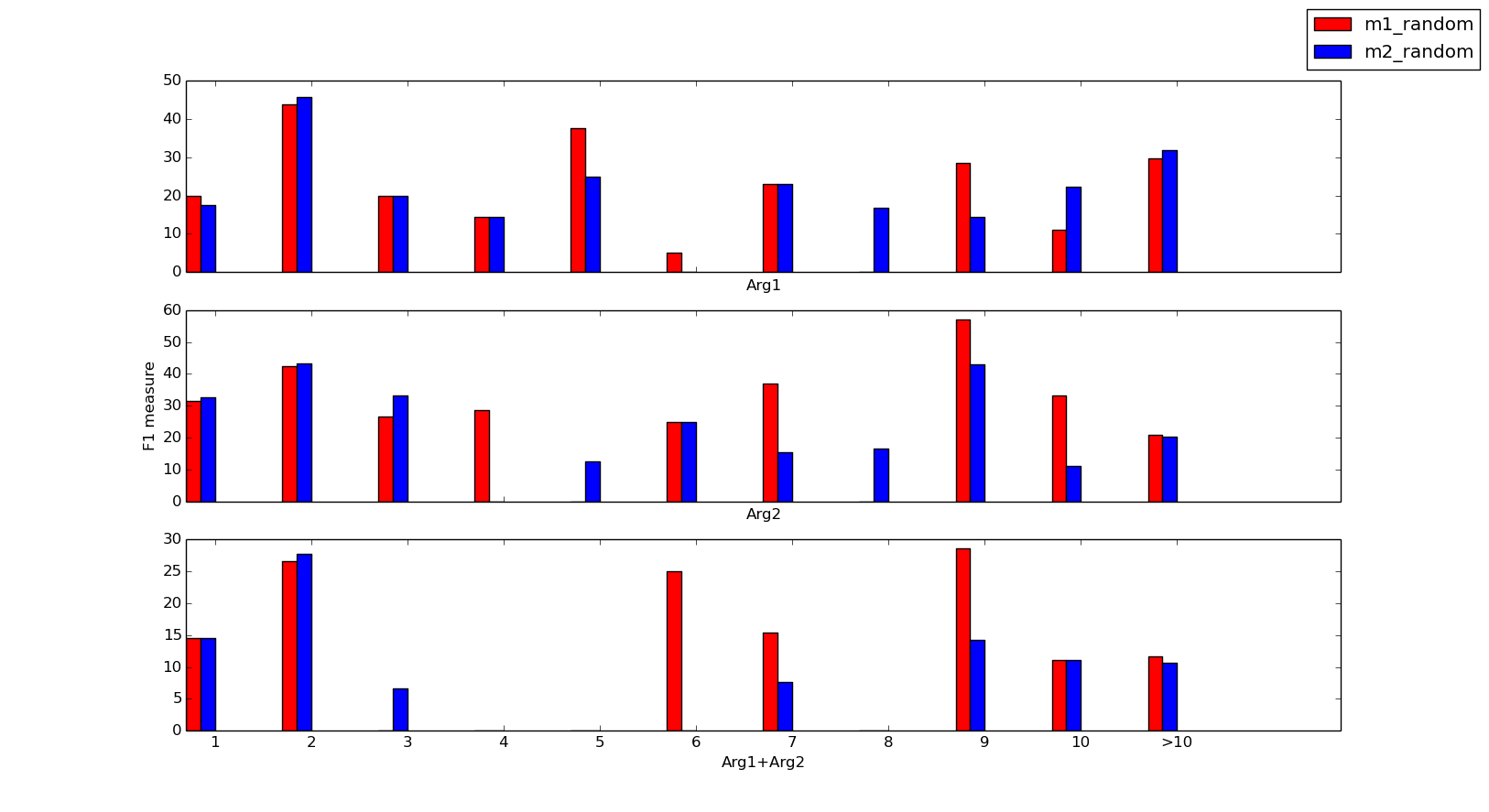}
        \caption{Plot of the distance-based F1 scores for Arg1 (top), Arg2 (middle) and Arg1+Arg2 (bottom)}
        \label{fig:distance-based-scores}
    \end{figure*}
    
    To evaluate our approach, we used the official CoNLL scoring module\footnote{\label{note2} available at https://github.com/attapol/conll16st} and modified it to calculate the performance for explicit relations only. Specifically, the scoring module provides the scores for the exact match for \texttt{Arg1} only, \texttt{Arg2} only and \texttt{Arg1+Arg2}, for every instance in the test set.
        
    For both models, the performance was evaluated at every epoch for a total of 50 evaluation points. Figure~\ref{fig:args} shows the F1 scores of the models for \texttt{Arg1} only, \texttt{Arg2} only and \texttt{Arg1+Arg2}. As the graphs show, after about 10 epochs all models seem to stabilize and learn at a much slower rate hence reaching a saturation point. 
    
    Table~\ref{tab:scores} shows the performance of our approaches compared to the state of the art systems. As the table shows, hand-engineered approaches still out perform our LSTM methods with F-measures between 55\% to 46\% for both \texttt{Arg1+Arg2}. However, compared to \citep{wang2015dcu}, both \texttt{m1} and \texttt{m2} outperform their RNN approach which did incorporate some hand-engineered features. It is also worthwhile to note that pre-computed embeddings result in slightly higher F1 measures for \texttt{Arg1+Arg2} than the random embeddings (25.75\% versus 23.75\%  for {\tt m2} and 24.89\% versus 22.75\% for {\tt m2}) . This is most likely because of the sparsity of the words used in the dataset. Since not all words are equally weighted, the system is unable to learn the relationship of those words in a given argument directly from the dataset. Therefore, having pre-computed embeddings assist in optimizing the learning process for those words.
 
   \begin{table*}[t]
    \centering
    \caption{F1 scores of our LSTM models for explicit relations compared to the best (hand-crafted) approaches and to \citep{wang2015dcu}}
    \begingroup
    \setlength{\tabcolsep}{4pt}
    \scalebox{0.9}{
        \begin{tabular}{|l|r|r|r|l|}
            \hline
             \textbf{Model}  &  \textbf{\texttt{Arg1+Arg2}}  &  \textbf{\texttt{Arg1}}  &  \textbf{\texttt{Arg2}}  &  \textbf{Method} \\ \hline
             \cite{wang2016two}    &  55.11\%  &  62.01\%  &  81.26\%  &  Linear classification \\
             \cite{schenk2016we}   &  54.41\%  &  61.97\%  &  78.87\%  &  CRF \\ 
             \cite{qin2016shallow}  &  53.44\%  &  60.99\%  &  79.94\%  &  SVM \\ 
             \cite{oepen2016opt}   &  51.37\%  &  60.72\%  &  75.83\%  &  SVM \\
             \cite{kong2016sonlp}  &  46.37\%  &  52.89\%  &  74.81\%  &  MaxEnt \\ 
              \texttt{m2\_GloVe}           &  25.75\%  &  42.06\%  &  41.49\%  &  Bidirectional LSTM \\
              \texttt{m1\_GloVe}           &  24.89\%  &  42.35\%  &  39.48\%  &  Bidirectional LSTM \\
              \texttt{m2\_random}          &  23.75\%  &  39.63\%  &  37.34\%  &  Bidirectional LSTM \\
              \texttt{m1\_random}          &  22.75\%  &  36.62\%  &  38.63\%  &  Bidirectional LSTM \\
             \cite{wang2015dcu}    &  20.52\%  &  28.55\%  & 41.78\% &  RNN \\ \hline
%            \citep{clac}                & 56.71\% & 75.95\% & 48.67\% & CRF \\ 
%            stepanov            & 55.66 & 79.07 & 49.36 \\ 
%            Soochow             & 51 & 73.98 & 42.7 \\ 
%            VTNLPS16            & 57.9 & 73.77 & 50.26 \\ 
%            nguyenlab           & 60.72 & 72.31 & 52.33 \\ 
%            rival2710           & 50.55 & 78.89 & 45.29 \\ 
%            devanshu            & 41.89 & 54.39 & 34.97 \\ 
%            nikko               & 40.76 & 52.3 & 32.82 \\ 
%            iitbhu              & 31.21 & 27.91 & 11.7 \\ \hline
        \end{tabular}
        }
    \endgroup
    \label{tab:scores}
    \end{table*}
    
    Because \texttt{Arg2} is structurally bound to the connective, in the case of explicit relations, identifying the connective gives strong evidence to locate \texttt{Arg2}. On the other hand, \texttt{Arg1} is much harder to identify as it can be located in various positions relative to \texttt{Arg2}. In the case of our LSTM based approach, it is interesting to note that while the F1 scores of \texttt{Arg1} and \texttt{Arg2} independently are quite lower than the state of the art, this difference diminishes drastically for the combined \texttt{Arg1+Arg2} F1 scores. This is because the neural network optimizes over an entire instance and hence tries to maximize the score for both arguments combined as opposed to independently optimizing \texttt{Arg1} and \texttt{Arg2} labeling. 
    
    To verify how our LSTM based approach handled long term dependencies, we separated the test dataset by distance and computed a distance-based F1 score. Recall from Section \ref{sec:related-work} that the distance is measured by the number of words between the closest words of \texttt{Arg1} and \texttt{Arg2} excluding the connective. Thus we count from the end of \texttt{Arg1} to the start of \texttt{Arg2} or the connective whichever comes first when \texttt{Arg1} precedes \texttt{Arg2} and from the end of \texttt{Arg2} or the connective whichever comes last to the start of \texttt{Arg1} when \texttt{Arg2} precedes \texttt{Arg1}. Figure~\ref{fig:distance-based-scores} shows the F1 scores for both the models learned with randomly initialized embeddings, calculated at their last epoch, as a function of the distance between \texttt{Arg1} and \texttt{Arg2}.  It is interesting to note that while model \texttt{m1\_random} seems to perform better on the longer distance based relations (greater than 9), model \texttt{m2\_random} still gets a better \texttt{Arg1+Arg2} accuracy score. Moreover, both models do not show any correlation in their F1 measure as the distance increases. This indicates that both models are unaffected by long and short distances between the arguments of a discourse relation.

\section{Conclusion and Future Work}
\label{sec:conclusion}
    This paper has presented a novel approach for argument labeling based on LSTMs. To our knowledge, this is the first attempt at using Deep Learning for this task that achieves good results without any features in comparison to the existing systems which rely entirely or partially on hand-crafted features. The approach adds value to this domain by decoupling the feature specificity of the PDTB dataset with the problem at hand. Thus, by using LSTM networks, it is possible to generalize the accuracy over different dataset. However, further research is required to prove this hypothesis >>. We experimented with two configurations of our model and showed that using the PDTB training set, our best model achieved 23.05\% F1 measure without feature engineering. We have shown that our LSTM-based models deal well with the long term dependencies of explicit discourse relations, an important problem with standard machine learning techniques.
    
    A number of improvements can be suggested over this approach. As shown in Table~\ref{tab:scores}, at this point feature-engineered approaches still provide a better performance than our LSTM-based method especially for labeling \texttt{Arg2}. To address this, it would be interesting to explore the use of a cascading network where a first network identifies the relative location of \texttt{Arg1} with respect to \texttt{Arg2} then forwards the discourse to a specialized network capable of learning only that specific type of location. In order to see how much weight is assigned to the discourse connective by the model, one could also test this approach on the implicit relations of the PDTB dataset. Finally, applying this approach on the Chinese Discourse Tree Dataset \citep{zhou2015chinese} would also be helpful in providing stronger evidence that the features that are learned are independent of the context and language.

\section*{Acknowledgement}
The authors would like to thank the anonymous reviewers for their feedback on an earlier version of the paper. This work was financially supported by the Natural Sciences and Engineering Research Council of Canada (NSERC).

%\section*{Acknowledgments}

%The acknowledgments should go immediately before the references.  Do not number the acknowledgments section. Do not include this section when submitting your paper for review.

% include your own bib file like this:
%\bibliographystyle{acl}
%\bibliography{acl2017}

\bibliographystyle{acl_natbib}

\appendix

\end{document}